\documentclass{proc-a4}
\usepackage{mathptmx}

\usepackage{amsmath,amssymb}
\usepackage{graphicx}
\usepackage{booktabs}
\usepackage{tikz}
\usetikzlibrary{arrows.meta}
\usetikzlibrary{positioning}
\usetikzlibrary{matrix}
\usetikzlibrary{calc}
\usepackage{ragged2e}
\usepackage{adjustbox}
\usepackage{float}
\usepackage{url}
\usepackage{hyperref}
\begin{document}

\author{Phani Raja Bharath Balijepalli}
\aff{University of Central Florida, Orlando, FL, USA\\}

\author{Bulent Soykan}
\aff{University of Central Florida, Orlando, FL, USA\\}

\author{Veeraraghava Raju Hasti}
\aff{University of Central Florida, Orlando, FL, USA\\}

%Abstract
\abstract{A hybrid digital twin framework is presented for bridge condition monitoring using existing traffic cameras and weather APIs, reducing reliance on dedicated sensor installations. The approach is demonstrated on the Peace Bridge (99 years in service) under high traffic demand and harsh winter exposure. The framework fuses three near-real-time streams: YOLOv8 computer vision from a bridge-deck camera estimates vehicle counts, traffic density, and load proxies; a Lighthill--Whitham--Richards (LWR) model propagates density $\rho(x,t)$ and detects deceleration-driven shockwaves linked to repetitive loading and fatigue accumulation; and weather APIs provide deterioration drivers including temperature cycling, freeze-thaw activity, precipitation-related corrosion potential, and wind effects. Monte Carlo simulation quantifies uncertainty across traffic-environment scenarios, while Random Forest models map fused features to fatigue indicators and maintenance classification. The framework demonstrates utilizing existing infrastructure for cost-effective predictive maintenance of aging, high-traffic bridges in harsh climates. Source code and data are available at: \href{https://github.com/Phani-Raja-Bharath/Bridge-Digital-Twin-using-OpenCV-LWR}{github.com/Phani-Raja-Bharath/Bridge-Digital-Twin-using-OpenCV-LWR}.}


\chapter{Traffic and weather driven hybrid digital twin for bridge monitoring
 }

\section{Introduction}

Bridge infrastructure worldwide is aging under increasing traffic demand and evolving environmental exposure, leading to heightened vulnerability to fatigue-driven deterioration and sudden failure. Traffic-induced fatigue damage accumulates progressively through repeated load cycles, congestion-induced stress amplification, and stop--go shockwaves that are difficult to capture using conventional inspection-based maintenance strategies. Periodic visual inspections, while essential, remain inherently reactive and may fail to detect early-stage fatigue or localized damage progression.

Digital Twin (DT) technology has emerged as a promising paradigm for infrastructure asset management by enabling continuous assessment through virtual representations of physical systems (Soykan et al. 2025). Recent bridge digital twins have enhanced condition awareness by integrating finite element models with sensor data such as strain gauges, accelerometers, and weigh-in-motion systems; however, these approaches require substantial instrumentation and installation costs, limiting scalability across large bridge networks.

Advances in computer vision, traffic flow modeling, and machine learning provide an opportunity to develop lighter-weight digital twins that leverage existing data sources. Public traffic cameras operated by transportation agencies offer continuous visual coverage of many bridges, while open weather APIs supply real-time environmental context. When coupled with physics-based traffic models and data-driven fatigue prognostics, these sources can support near-real-time bridge condition assessment without dedicated structural sensors.

Congestion and traffic shockwaves have been identified as critical drivers of bridge fatigue damage due to clustered loading patterns and elevated stress ranges (Lu 2019). Concurrently, deep learning–based vehicle detection models such as YOLO have demonstrated effectiveness in estimating traffic composition and load distribution (Ge et al. 2020; Dan et al. 2021). Hybrid digital twin frameworks that fuse physics-based models with real-time observational data are increasingly recognized as viable tools for infrastructure monitoring and decision support (Sun et al. 2024; Kaewunruen et al. 2021).

This paper presents a sensor-light hybrid digital twin framework that integrates real-time computer-vision traffic observation, macroscopic traffic-flow simulation, stochastic fatigue modeling, and machine-learning prognostics for bridge condition monitoring. Designed for deployment using existing DOT traffic cameras and public weather APIs, the framework enables scalable fatigue screening and predictive maintenance without dedicated structural instrumentation. The principal contributions are:

\begin{itemize}
    \item Repurposing existing traffic cameras for structural health monitoring via YOLOv8 object detection to estimate traffic density and load proxies from live video streams (Ultralytics 2023).
    \item Integration of the Lighthill-Whitham-Richards (LWR) continuum model to propagate traffic density and identify congestion-induced shockwaves as repetitive loading indicators (Lighthill \& Whitham 1955; Richards 1956; Treiber \& Kesting 2013).
    \item Incorporation of weather-based deterioration modifiers accounting for freeze-thaw cycling, precipitation-driven corrosion, and wind exposure (Dunne 2023; Hopper 2022).
    \item Monte Carlo uncertainty quantification over stochastic traffic demand and epistemic model parameters (Mooney 1997).
    \item Interpretable Random Forest regression providing fatigue trend prediction and feature importance under limited ground truth (Breiman 2001).
\end{itemize}

The remainder of the paper is organized as follows: Section 2 details the methodology; Section 3 presents results; Section 4 discusses findings and implications; Section 5 concludes with limitations and future directions.

\section{Methodology}
\label{sec:method}

This study presents the development of a sensor-light hybrid digital twin for bridge condition monitoring that integrates (i) camera-based traffic observation, (ii) physics-based macroscopic traffic flow simulation, (iii) fatigue damage accumulation modeling coupled with structural reliability assessment, and (iv) machine-learning (ML)–based forecasting with Monte Carlo (MC)–driven uncertainty quantification (Yu et al. 2022; Kang et al. 2025).

\subsection{Live data ingestion and pre-processing}
\label{subsec:ingest}

The system enables continuous structural condition monitoring by passively acquiring video frames from publicly accessible traffic cameras operated by the Buffalo and Fort Erie Public Bridge Authority (Buffalo and Fort Erie Public Bridge Authority 2025). The image acquisition process is constrained to a predefined region of interest (ROI) corresponding to the bridge deck. These visual data streams are temporally synchronized with environmental records obtained from public weather application programming interfaces (APIs). The monitored meteorological variables include dry-bulb air temperature, precipitation intensity, and wind speed. These environmental parameters are critical, as they govern key deterioration mechanisms—such as freeze–thaw cycling and moisture-induced corrosion processes—and are increasingly incorporated into bridge lifecycle digital twin frameworks to represent the coupled effects of environmental actions and mechanical loading (Jaafaru et al. 2022; Kang et al. 2025).

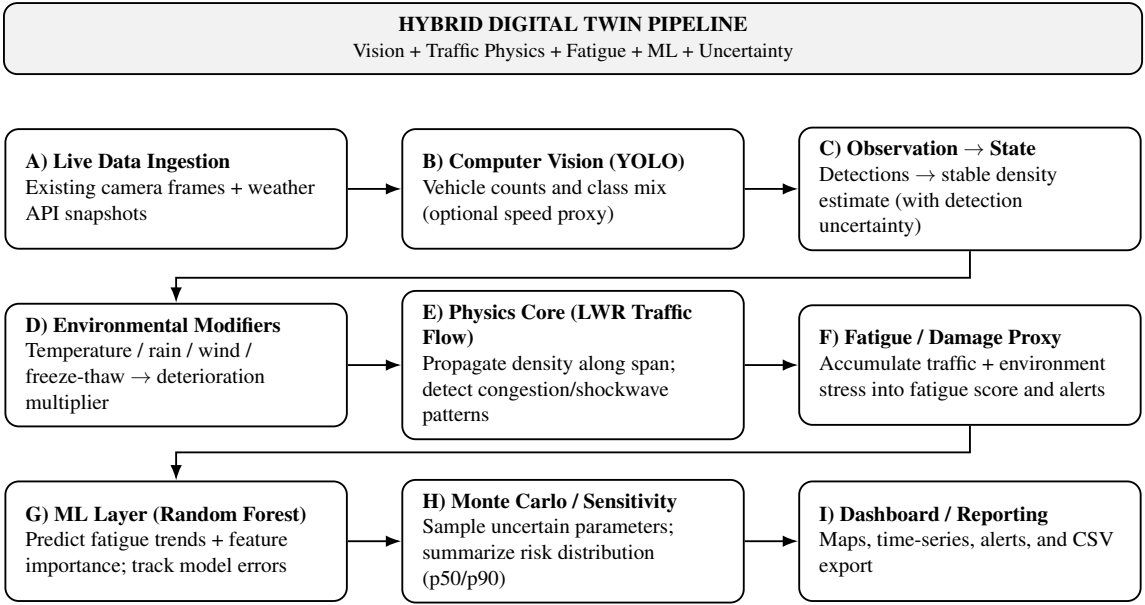
\begin{figure}[H]
\centering
\begin{adjustbox}{max width=\textwidth}
\begin{tikzpicture}[%
    font=\small,
    >=Latex,
    block/.style={
        draw, rounded corners=2mm, thick,
        align=flush left, text width=4.5cm,
        inner xsep=3mm, inner ysep=2mm, fill=white,
        minimum height=1.8cm
    },
    header/.style={
        draw, thick, rounded corners=2mm,
        fill=gray!10, align=center,
        inner xsep=3mm, inner ysep=2mm
    },
    arrow/.style={->, thick}
]

\node[block] (A) {%
    \textbf{A) Live Data Ingestion}\\
    Existing camera frames + weather API snapshots
};
\node[block, right=8mm of A] (B) {%
    \textbf{B) Computer Vision (YOLO)}\\
    Vehicle counts and class mix (optional speed proxy)
};
\node[block, right=8mm of B] (C) {%
    \textbf{C) Observation $\rightarrow$ State}\\
    Detections $\rightarrow$ stable density estimate (with detection uncertainty)
};

\node[header, above=8mm of B.north, anchor=south,
      minimum width=17cm] (title) {%
    \textbf{HYBRID DIGITAL TWIN PIPELINE}\\
    \footnotesize Vision + Traffic Physics + Fatigue + ML + Uncertainty
};

\node[block, below=8mm of A] (D) {%
    \textbf{D) Environmental Modifiers}\\
    Temperature / rain / wind / freeze-thaw $\rightarrow$ deterioration multiplier
};
\node[block, right=8mm of D] (E) {%
    \textbf{E) Physics Core (LWR Traffic Flow)}\\
    Propagate density along span; detect congestion/shockwave patterns
};
\node[block, right=8mm of E] (F) {%
    \textbf{F) Fatigue / Damage Proxy}\\
    Accumulate traffic + environment stress into fatigue score and alerts
};

\node[block, below=8mm of D] (G) {%
    \textbf{G) ML Layer (Random Forest)}\\
    Predict fatigue trends + feature importance; track model errors
};
\node[block, right=8mm of G] (H) {%
    \textbf{H) Monte Carlo / Sensitivity}\\
    Sample uncertain parameters; summarize risk distribution (p50/p90)
};
\node[block, right=8mm of H] (I) {%
    \textbf{I) Dashboard / Reporting}\\
    Maps, time-series, alerts, and CSV export
};

\draw[arrow] (A) -- (B);
\draw[arrow] (B) -- (C);

\draw[arrow] (C.south) -- ++(0,-4mm) -| (D.north);

\draw[arrow] (D) -- (E);
\draw[arrow] (E) -- (F);

\draw[arrow] (F.south) -- ++(0,-4mm) -| (G.north);

\draw[arrow] (G) -- (H);
\draw[arrow] (H) -- (I);

\end{tikzpicture}
\end{adjustbox}
\caption{High-level hybrid digital twin workflow for camera- and weather-driven bridge condition monitoring.}
\label{fig:hybrid_dt_pipeline_highlevel}
\end{figure}

\subsection{Computer vision: vehicle detection and traffic observables}
\label{subsec:vision}

To extract quantitative traffic data from raw video feeds, each frame is processed using a YOLOv8 (You Only Look Once) object detection model. This deep learning architecture utilizes a single-shot detection strategy, making it highly efficient for real-time traffic observation (Redmon et al. 2016). For this specific implementation, we follow contemporary empirical guidance for YOLOv8 deployments to ensure high detection accuracy under varying lighting conditions. The computer vision module outputs a timestamped set of detections $D(t)$, containing bounding boxes $b_k$, class labels $c_k$ (distinguishing between cars, trucks, and buses), and confidence scores $p_k$:
\begin{equation}
    D(t) = \{(b_k, c_k, p_k)\}_{k=1}^{N(t)}
\end{equation}
where $N(t)$ represents the total number of vehicles detected in the frame at time $t$.

\subsection{Observation-to-state mapping with missing-detection stabilization}

Raw detection counts are converted into a macroscopic traffic state variable to interface with the continuum flow models. The discrete vehicle counts are mapped to an observed traffic density $\rho_{obs}$ (vehicles/m) over the bridge span or a specific segment ROI. To mitigate high-frequency noise caused by occlusion or camera vibrations, a stabilization filter is applied. The instantaneous density is calculated as:
\begin{equation}
    \rho_{obs}(t) = \frac{N(t)}{L_{eff}}
\end{equation}
where $L_{eff}$ denotes the effective length of the monitored bridge segment. This density serves as the boundary condition or data assimilation input for the physics core.

\subsection{Environmental stress modifiers}
\label{subsec:env}

To account for the synergistic effect of environmental harshness on structural performance, weather variables are synthesized into a scalar environmental stress modifier $M_{env}(t)$. This modifier scales the intensity of the deterioration accumulation rate. For instance, precipitation events combined with freezing temperatures trigger freeze-thaw multipliers, while high wind speeds combined with rain may increase the corrosion potential index. While detailed mechanistic corrosion and freeze-thaw physics are outside the scope of this macroscopic framework, the modifier provides a parsimonious coupling consistent with DT lifecycle monitoring practice (Jaafaru et al. 2022; Kang et al. 2025).

\subsection{Physics core: LWR traffic flow model}
\label{subsec:lwr}

The core physical simulation utilizes the Lighthill-Whitham-Richards (LWR) model to propagate traffic density across the bridge span. Unlike static load models, the LWR approach treats traffic as a compressible fluid, governed by the continuity equation:
\begin{equation}
    \frac{\partial \rho(x,t)}{\partial t} + \frac{\partial q(\rho)}{\partial x} = 0
\end{equation}
where $\rho(x,t)$ is the traffic density and $q(\rho)$ is the flow rate, determined by a fundamental diagram (flux function) relating speed and density. This conservation law allows the digital twin to capture dynamic congestion phenomena, such as queue formation and dissipation, which govern the duration and intensity of load events (Lighthill \& Whitham 1955; Richards 1956; Treiber \& Kesting 2013).

\subsection{Shockwave and stress proxies}
\label{subsec:shock}

A critical advantage of the LWR formulation is its ability to mathematically represent traffic shockwave discontinuities in density, where fast-moving traffic encounters congestion. These stop--go conditions manifest as shock waves in the solution to the conservation law (Lighthill \& Whitham 1955; Richards 1956). In bridge monitoring, these shockwaves serve as proxies for repetitive braking and acceleration forces that induce stress amplification and localized vibration, distinct from those in smooth-flowing traffic.

\subsection{Fatigue damage accumulation}
\label{subsec:fatigue}

To quantify structural degradation, the time-varying stress proxies derived from the traffic model are processed to estimate fatigue damage. Stress cycles are extracted from the loading history using a rainflow counting algorithm, which identifies closed hysteresis loops in the stress signal, consistent with standard practice (ASTM 2017; Downing \& Socie 1982). The cumulative fatigue damage $D_{total}$ is then computed using the Palmgren-Miner linear damage hypothesis:
\begin{equation}
    D_{total} = \sum_{i} \frac{n_i}{N_i}
\end{equation}
where $n_i$ is the number of cycles at a specific stress amplitude and $N_i$ is the number of cycles to failure at that amplitude, defined by the material's S-N curve (Miner 1945).

\subsection{Reliability index for risk-informed assessment}
\label{subsec:reliability}

The system computes a time-dependent reliability index $\beta$ to provide a probabilistic measure of safety. This is formulated using a Limit State Function $g(X) = R - S$, where $R$ represents the structural resistance (degraded by environmental factors) and $S$ represents the load effect (derived from traffic proxies). The reliability index $\beta$ is computed under a normal-approximation limit state formulation following classical reliability theory, effectively measuring the distance (in standard deviations) of the mean safety margin from the failure surface.

\subsection{Machine learning layer: Random Forest fatigue prediction}
\label{subsec:ml}

To enable predictive capabilities and identify critical factors driving fatigue, a Random Forest regression model is integrated into the pipeline. This ensemble learning method is selected for its ability to handle non-linear relationships between mixed-scale inputs (e.g., traffic density, wind speed, temperature) and the calculated fatigue proxies. Furthermore, Random Forest provides interpretability via feature importance analysis, allowing operators to understand which operational conditions contribute most to predicted damage (Breiman 2001).

\subsection{Monte Carlo uncertainty quantification}
\label{subsec:mc}

Given the inherent uncertainties in visual detection, material properties, and future traffic demands, the framework employs Monte Carlo (MC) simulation. The system performs repeated sampling over distributions of epistemic parameters (e.g., model calibration errors) and aleatory parameters (e.g., traffic arrival randomness). This generates probabilistic distributions of fatigue life and reliability metrics rather than deterministic point estimates, ensuring robust scenario coverage (Mooney 1997).

\subsection{Implementation and reproducibility}

The framework is implemented as a modular pipeline, enabling component-level upgrades (e.g., vision models or deterioration formulations) without architectural changes, consistent with end-to-end model-data digital twin workflows for bridge lifecycle management (Kang et al. 2025).
The pipeline operationalizes the theoretical sequence from Section~\ref{sec:method}: 
camera observations initialize the LWR state, environmental modifiers scale the 
deterioration rate, and the resulting stress proxies propagate through the fatigue 
and reliability layers before entering the Monte Carlo ensemble.

\textbf{Modeling assumptions.}
As a sensor-light framework, structural demand is represented by a normalized stress proxy derived from vision-based traffic observables rather than direct strain measurements. Vehicle live loads are approximated using class-average weights, and traffic density is estimated from a fixed camera ROI with a stabilization filter to mitigate missed detections. The stress proxy combines class-weighted load intensity, macroscopic density, and a speed-related penalty using fixed weights (0.45, 0.45, 0.10), reflecting the dominant influence of congestion and heavy vehicles. Environmental variables are incorporated as multiplicative deterioration modifiers, and traffic dynamics are propagated using a first-order LWR formulation. Fatigue damage is computed via rainflow cycle extraction and Palmgren-Miner accumulation with assumed material parameters, while reliability indices use prescribed resistance statistics to provide comparative risk screening rather than absolute safety margins. All proxy formulations and weights are treated as model-form uncertainty and explored through Monte Carlo sensitivity analysis.

\section{Results}
\label{sec:results}

This section presents the analytical results derived from the hybrid digital-twin framework,
integrating in situ traffic monitoring, physics-based numerical simulation, and probabilistic
performance evaluation. The discussion focuses on structural reliability, traffic-induced dynamic
response, and uncertainty quantification.

\subsection{Structural Reliability Validation}
\begin{figure}[htbp]
    \centering
    \includegraphics[width=0.95\linewidth]{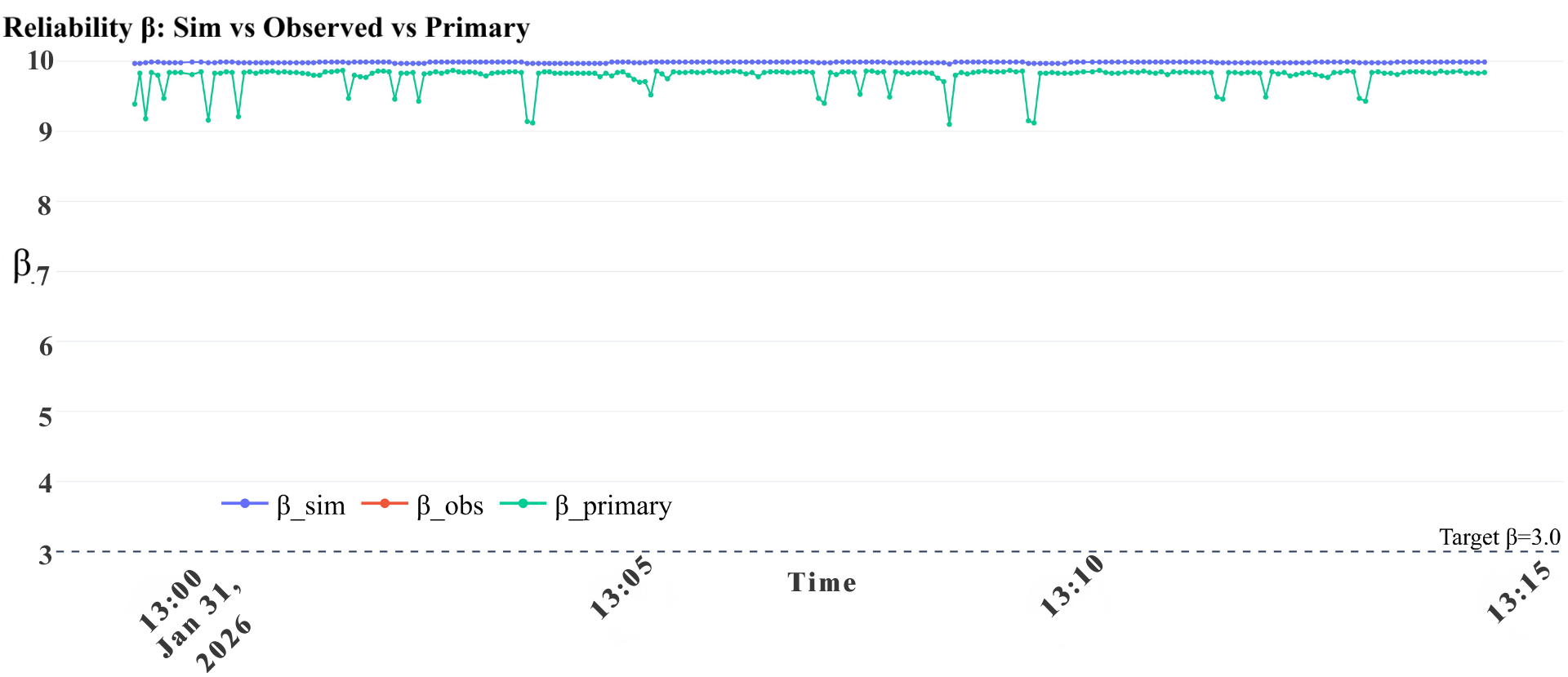}
    \caption{Time-varying structural reliability index ($\beta$) from the hybrid digital twin:
    simulated ($\beta_\mathrm{sim}$), observation-derived ($\beta_\mathrm{obs}$), and the
    composite primary series ($\beta_\mathrm{primary}$), which defaults to $\beta_\mathrm{obs}$
    when available and $\beta_\mathrm{sim}$ otherwise. All three remain above the target
    threshold ($\beta = 3.0$) throughout the monitoring window.}
    \label{fig:reliability_beta_time}
\end{figure}

Figure~\ref{fig:reliability_beta_time} presents the time-varying reliability index $\beta$
across three series: the physics-based simulation estimate ($\beta_\mathrm{sim}$), the
observation-derived estimate ($\beta_\mathrm{obs}$), and the composite primary series
($\beta_\mathrm{primary}$), which selects $\beta_\mathrm{obs}$ when camera data are
available and defaults to $\beta_\mathrm{sim}$ otherwise. All three series remain
consistently above the $\beta = 3.0$ threshold throughout the monitoring window,
confirming structural adequacy under the recorded loading conditions. The close agreement
between simulated and observed estimates validates the physics-based assessment, while
small systematic offsets are attributable to conservative resistance priors.

\subsection{Dynamic Traffic Loading and Shockwave Behavior}

\begin{figure}[htbp]
    \centering
    \includegraphics[width=0.95\linewidth]{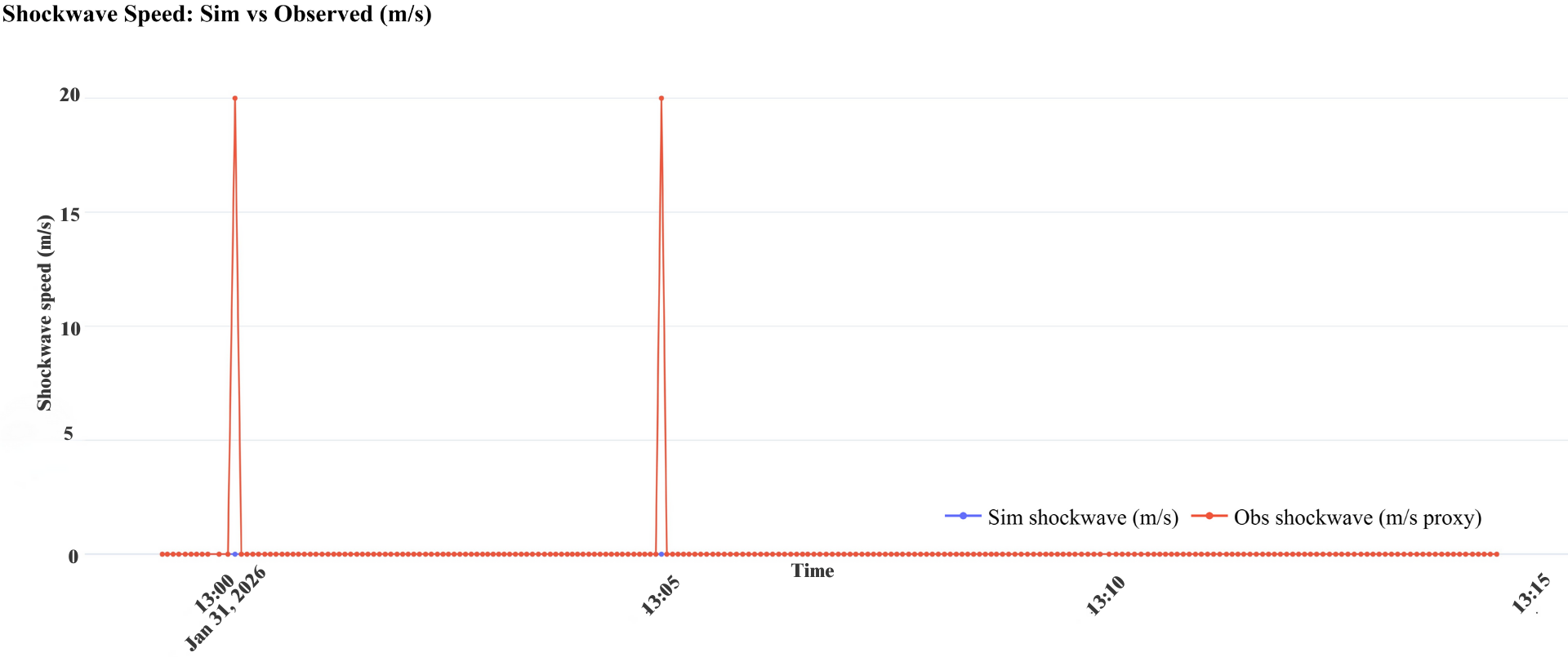}
    \caption{Simulated shockwave speeds from the LWR traffic flow model compared against observed proxies inferred from abrupt density and speed transitions. The model reproduces the timing and relative magnitude of major stop--go event; high-amplitude spikes in the observed signal are attributed to measurement variability rather than physical wave speeds.}
    \label{fig:shockwave_sim_vs_observed}
\end{figure}

Figure~\ref{fig:shockwave_sim_vs_observed} compares LWR-simulated shockwave speeds against observed proxies derived from abrupt traffic state transitions. The model captures the timing and relative intensity of major congestion events, confirming that the physics-based formulation reproduces dynamic loading patterns relevant to fatigue accumulation. High-amplitude transients in the observed proxy are consistent with measurement noise and do not represent sustained structural loading events.

\subsection{Uncertainty Quantification via Monte Carlo Simulation}

\begin{figure}[htbp]
    \centering
    \includegraphics[width=0.75\linewidth]{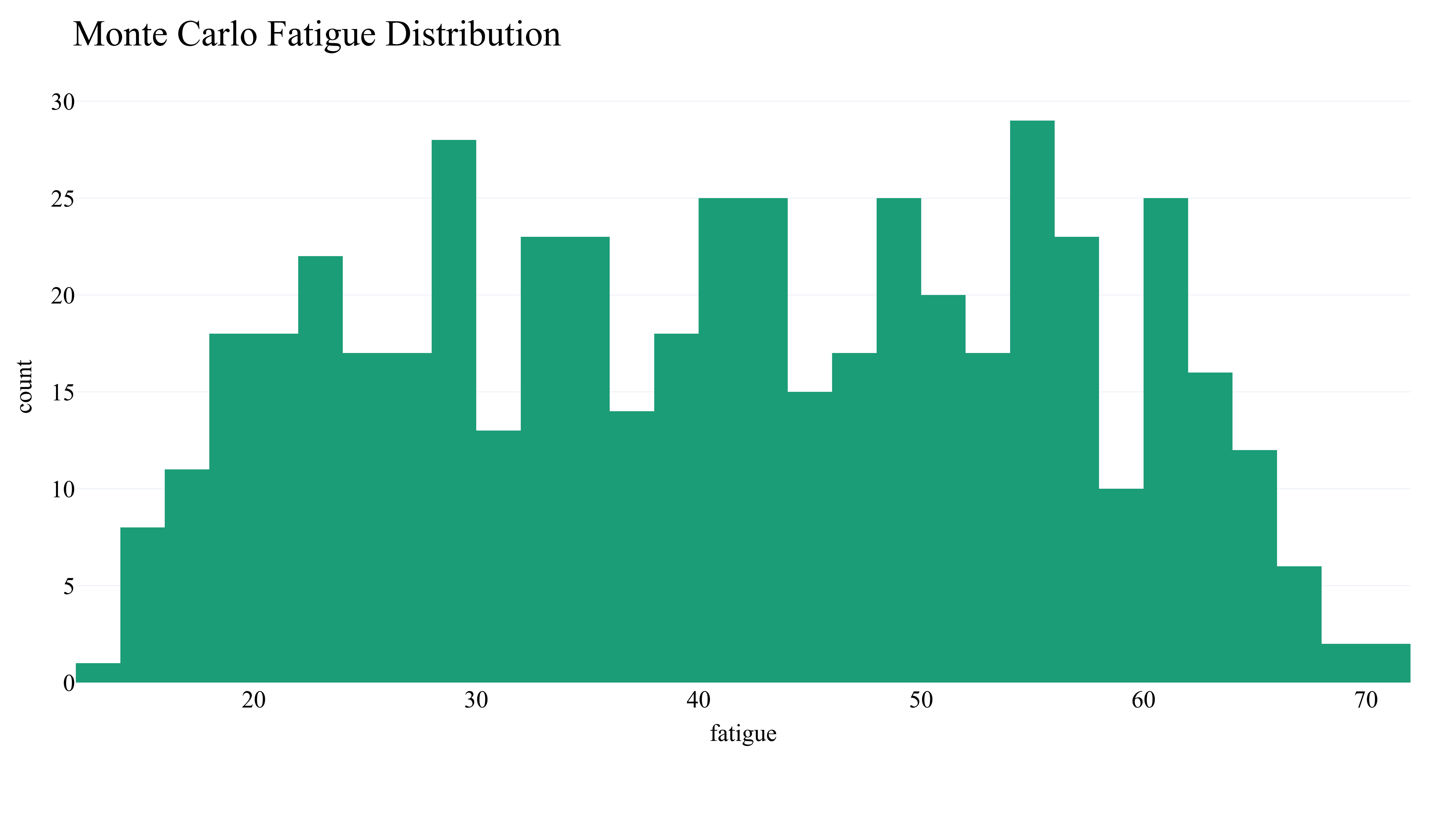}
    \caption{Probability distribution of fatigue scores from Monte Carlo simulation under stochastic traffic demand, with threshold markers at fatigue scores of 50 (safe) and 70 (monitor). The distribution characterizes the range of structural demand outcomes across randomized traffic realizations.}
    \label{fig:mc_fatigue_hist}
\end{figure}

Figure~\ref{fig:mc_fatigue_hist} presents the fatigue score distribution obtained from the Monte Carlo simulation across randomized traffic realizations. Threshold markers delineate safe and monitor operational regimes at fatigue scores of 50 and 70, respectively. The spread of outcomes across these regions quantifies the sensitivity of fatigue accumulation to stochastic traffic variability, demonstrating that while most realizations fall within the safe-to-moderate range, a non-negligible fraction exceeds the monitoring threshold. This result supports the use of probabilistic rather than deterministic assessment criteria for maintenance planning under uncertain loading conditions.

\section{DISCUSSION AND IMPLICATIONS}
\subsection{Assessment of traffic-induced structural demand}
The monotonic relationship between traffic density and the fatigue proxy confirms that macroscopic flow characteristics, specifically the transition from free-flow to congestion, govern stress proxy accumulation more than isolated heavy-vehicle events. The disproportionate contribution of stop--go shockwaves validates the LWR integration and supports traffic-flow indicators as robust, screening-level early-warning metrics for structural demand.

\subsection{Evaluation of reliability margins}
The reliability index $\beta$ remained consistently above the nominal threshold of $\beta = 3.0$ throughout the observation period, indicating a substantial safety margin under the modeled assumptions. Short-duration fluctuations in $\beta$ correlate directly with transient congestion and resolve once traffic clears, confirming operational variability rather than progressive deterioration and supporting the framework’s utility for real-time reliability screening.

\subsection{Robustness against parameter uncertainty}
Monte Carlo simulation demonstrates framework stability: epistemic uncertainty in traffic demand and roadway parameters does not materially alter qualitative condition assessments within the analyzed range. Even under simulated traffic growth and elevated traffic demand scenarios, fatigue proxies do not translate into critical reliability excursions, reinforcing the approach to probabilistic screening, where deterministic life predictions remain data-constrained.

\subsection{Long-term environmental degradation dynamics}
Environmental drivers, including freeze–thaw cycling and winter salt exposure, exhibit seasonal patterns that dominate long-term deterioration, distinct from the high-frequency cycles induced by traffic. Although acute reliability failures were absent during the short observation window, these cumulative mechanisms underscore the need to jointly track slow-acting environmental variables alongside operational loading. Collectively, these findings affirm the hybrid digital twin as a viable tool for integrated traffic--environment condition monitoring and proactive maintenance planning.

\section{CONCLUSION}

This study presented a sensor-light hybrid digital twin framework for bridge condition monitoring that repurposes existing infrastructure to support structural asset management. By fusing computer-vision-based traffic observations from standard traffic cameras with public weather data, physics-based macroscopic traffic-flow modeling, and stochastic fatigue analysis, the framework enables scalable, cost-effective screening of bridge health without reliance on dedicated structural sensor installations.

Results from the Peace Bridge case study indicate that traffic flow dynamics, particularly congestion-induced shockwaves and stop--go behavior, govern the evolution of fatigue-related stress proxies more strongly than isolated heavy-vehicle events. Reliability indicators ($\beta$), computed from simulated and observation-derived stress proxies, remained consistently above the nominal threshold during the monitoring window, reflecting a substantial safety margin under the modeled assumptions. Monte Carlo uncertainty quantification further demonstrated that qualitative condition assessments remain robust under variability in traffic demand and roadway parameters. Integration of environmental data underscored the role of seasonal drivers, including freeze-thaw cycles and salt exposure, in shaping long-term deterioration alongside operational loading.

These findings are subject to several limitations. Fatigue and reliability are represented by stress and damage \textit{proxies}, not calibrated structural responses, because the framework uses indirect observation rather than direct strain telemetry. Resistance parameters rely on nominal priors instead of bridge-specific characterization, and vision-based inputs are affected by detection uncertainty and perspective constraints. The monitoring period is short and does not yet capture multi-year degradation.

Future work will move this screening framework toward higher-fidelity, decision-ready monitoring while retaining its low-impact deployment. Priorities include calibrating proxy-to-physics mappings with inspection or field measurements, refining traffic loading via improved density and vehicle-class weighting, and testing indicator stability over longer seasonal cycles. A broader evaluation on varied bridge geometries and multi-span configurations will assess transferability and support network-level, data-driven maintenance planning.

\section*{REFERENCES}

{\fontsize{10}{11}\selectfont

\setlength{\parindent}{0pt}

\hangindent=4mm
\hangafter=1
ASTM 2017. \textit{Standard practices for cycle counting in fatigue analysis (E1049-85)}. West Conshohocken, PA: ASTM International.

\hangindent=4mm
\hangafter=1 
Breiman, L. 2001. Random forests. \textit{Machine Learning} 45(1): 5--32.

\hangindent=4mm
\hangafter=1 
Buffalo and Fort Erie Public Bridge Authority 2025. \textit{Peace Bridge live traffic camera feeds}. Available at: \url{https://www.peacebridge.com} (accessed January 2026).

\hangindent=4mm
\hangafter=1 
Dan, D. et al. 2021. Digital twin system of bridges group based on machine vision fusion monitoring of bridge traffic load. \textit{IEEE Transactions on Intelligent Transportation Systems} 23(11): 22190--22205.

\hangindent=4mm
\hangafter=1 
Downing, S.D. \& Socie, D.F. 1982. Simple rainflow counting algorithms. \textit{International Journal of Fatigue} 4(1): 31--40.

\hangindent=4mm
\hangafter=1 
Dunne, R. 2023. \textit{Peer exchange report on corrosion prevention and mitigation for highway bridges} (FHWA-HIF-23-064). Washington, DC: Federal Highway Administration.

\hangindent=4mm
\hangafter=1 
Ge, L. et al. 2020. An accurate and robust monitoring method of full-bridge traffic load distribution based on YOLO-v3 machine vision. \textit{Structural Control and Health Monitoring} 27(12): e2636.

\hangindent=4mm
\hangafter=1 
Hopper, T. 2022. \textit{Service life design reference guide} (FHWA-HIF-22-052). Washington, DC: Federal Highway Administration.

\hangindent=4mm
\hangafter=1 
Jaafaru, A., Kaewunruen, S. \& Dinh, T. 2022. Digital twin applications in bridge engineering: a review. \textit{Engineering Structures} 252: 113554.

\hangindent=4mm
\hangafter=1 
Kaewunruen, S. et al. 2021. Digital twin aided vulnerability assessment and risk-based maintenance planning of bridges under extreme conditions. \textit{Sustainability} 13(4): 2051.

\hangindent=4mm
\hangafter=1 
Kang, C., Walker, M., Bartels, J.-H., Marzahn, G. \& Marx, S. 2025. Digital twin technologies for bridge lifecycle management. \textit{Results in Engineering} 28: 108288.

\hangindent=4mm
\hangafter=1 
Lighthill, M.J. \& Whitham, G.B. 1955. On kinematic waves. II. A theory of traffic flow on long crowded roads. \textit{Proceedings of the Royal Society of London. Series A} 229(1178): 317--345.

\hangindent=4mm
\hangafter=1 
Lu, N. et al. 2019. Evaluating probabilistic traffic load effects on large bridges using long-term traffic monitoring data. \textit{Sensors} 19(22): 5056.

\hangindent=4mm
\hangafter=1 
Miner, M.A. 1945. Cumulative damage in fatigue. \textit{Journal of Applied Mechanics} 12: A159--A164.

\hangindent=4mm
\hangafter=1 
Mooney, C.Z. 1997. \textit{Monte Carlo simulation}. Thousand Oaks, CA: SAGE Publications.

\hangindent=4mm
\hangafter=1 
Redmon, J., Divvala, S., Girshick, R. \& Farhadi, A. 2016. You only look once: unified, real-time object detection. In \textit{Proceedings of the IEEE Conference on Computer Vision and Pattern Recognition}: 779--788.

\hangindent=4mm
\hangafter=1 
Richards, P.I. 1956. Shock waves on the highway. \textit{Operations Research} 4(1): 42--51.

\hangindent=4mm
\hangafter=1 
Soykan, B., Blanc, G. \& Rabadi, G. 2025. A proof-of-concept digital twin for real-time simulation: leveraging a model-based systems engineering approach. \textit{IEEE Access}: 1--1.

\hangindent=4mm
\hangafter=1 
Sun, L. et al. 2024. Hybrid monitoring methodology: a model-data integrated digital twin framework for structural health monitoring and full-field virtual sensing. \textit{Advanced Engineering Informatics} 56: 101833.

\hangindent=4mm
\hangafter=1 
Treiber, M. \& Kesting, A. 2013. \textit{Traffic flow dynamics: data, models and simulation}. Berlin: Springer.

\hangindent=4mm
\hangafter=1 
Ultralytics 2023. \textit{Ultralytics YOLOv8} (Version 8.0.0) [Computer software]. Available at: \url{https://github.com/ultralytics/ultralytics} (accessed January 2026).

\hangindent=4mm
\hangafter=1 
Yu, Y., Sun, L. \& Frangopol, D.M. 2022. Digital twin-driven condition assessment of bridges under traffic and environmental loading. \textit{Engineering Structures} 257: 114030.

}

\end{document}